\documentclass{article} 
\usepackage{iclr2021_conference,times}

\usepackage{amsmath,amsfonts,bm}

\def\eqref#1{equation~\ref{#1}}

\def\1{\bm{1}}

\DeclareMathAlphabet{\mathsfit}{\encodingdefault}{\sfdefault}{m}{sl}
\SetMathAlphabet{\mathsfit}{bold}{\encodingdefault}{\sfdefault}{bx}{n}

\usepackage{hyperref}
\usepackage{url}
\usepackage{listings}   
\usepackage{booktabs}
\usepackage[ruled,vlined]{algorithm2e}
\usepackage{graphicx}
\usepackage{caption}

\title{Planning with affordances: Integrating learned affordance models and symbolic planning}

\author{Rajesh Mangannavar\\
Oregon State University\\
Corvallis, OR 97330, USA \\
\texttt{\{mangannr\}@oregonstate.edu} \\
}

\newcommand{\bi}{\begin{itemize}}
\newcommand{\ei}{\end{itemize}}

\newcommand{\comment}[1]{}

\def\noteme#1{}
\def\notecoauth#1{\ }

\iclrfinalcopy 
\begin{document}
\maketitle
\begin{abstract}




Intelligent agents working in real-world environments must be able to learn about the environment and its capabilities which enable them to take actions to change to the state of the world to complete a complex multi-step task in a photorealistic environment. Learning about the environment is especially important to perform various multiple-step tasks without having to redefine an agent's action set for different tasks or environment settings. In our work, we augment an existing task and motion planning framework with learned affordance models of objects in the world to enable planning and executing multi-step tasks using learned models. Each task can be seen as changing the current state of the world to a given goal state. The affordance models provide us with what actions are possible and how to perform those actions in any given state. A symbolic planning algorithm uses this information and the starting and goal state to create a feasible plan to reach the desired goal state to complete a given task. We demonstrate our approach in a virtual 3D photorealistic environment,  AI2-Thor, and evaluate it on real-world tasks. Our results show that our agent quickly learns how to interact with the environment and is well prepared to perform tasks such as "Moving an object out of the way to reach the desired location."
\end{abstract}

\section{Introduction}



In real-world environments, the ability to come up with multi-step plans for a particular task is an important skill for an intelligent agent. Moreover, it is equally important that the agent be able to interact with the environment to execute this plan. For example, for efficient navigation of an environment, the agent must generate multi-step plans, including navigating through different rooms while clearing out any objects blocking the path. However, the agent must also be able to interact with the environment correctly for actions such as opening the door or picking up an object that is blocking the way. Here the agent needs to make both discrete decisions (picking up an object) as well as continuous decisions (the choice of its exact position/orientation to be able to pick up the object).  


The aforementioned problem falls under a class of problems called the task and motion planning (TAMP) problems. This class of problems generally deals with environments that contain several objects, and the agent is expected to navigate in the environment, interact and change the state of objects to complete a given task.

The more specific problem we would like our agent to solve is, to be able to be dropped into a real-world environment or a 3D virtual environment and given a task to perform, learn about the objects in the world, produce and execute plans which achieve complex multi-step tasks by making the required discrete and continuous decisions.

To solve the defined TAMP problem, we need a system that is capable of learning affordances in the world in a way that enables composing these affordance models by a higher-level symbolic planning algorithm to solve a given task. The complexity of the problem is increased further when we want to solve the tasks in a hybrid discrete and continuous world. Hence, we need a framework that enables us to learn and plan on flexible affordance models of the world and allows us to plan in the hybrid discrete and continuous world.

A lot of existing work on embodied agents performing tasks in real-world domains approach the problem of doing tasks using a pure learning approach where a reinforcement learning (RL) agent is trained which receives a reward from the world upon completing the required tasks. However, the approaches generally using learning either focus on learning exploration policies (\cite{Learning_Exploration_policies_for_Navigation})  or learn policies for specific tasks  (\cite{Cognitive_Mapping_and_Planning_for_Visual_Navigation}) which fail to generalize to newer or more complex tasks very well. 

Some of the more recent works use affordance learning with reinforcement to complete tasks in a given environment. The affordance models learn to predict possible actions in the given state of the world and are used to train an RL agent using these models (
 \cite{Learning_Affordance_Landscapes_Interaction_Exploration}, \cite{Learning_About_Objects_by_Learning_to_Interact}).
Other approaches to using model-based RL to learn a dynamics model or a transition model of the world and use this to train an RL agent to perform tasks (\cite{Visual_Semantic_Planning_Using_Deep_Successor_Representations}, \cite{Theory_of_Affordances_in_Reinforcement_Learning}). These approaches also have the drawback of not being able to complete a diverse set of multi-step complex tasks as they enable learning about what actions are possible in a given environment but not how to perform those actions successfully. 


Planning-based approaches to solving TAMP problems in hybrid discrete and continuous domains have been tried with some success (\cite{Using_classical_planners_for_tasks}).PDDLStream (\cite{PDDLStream_main}) is one of the more recent frameworks in which a hybrid discrete and continuous state space is reduced to an common planning problem by using conditional samplers. The process of learning these samplers has been added to PDDLStream in \cite{Learning_compositional_models} for more effective sampling. These approaches have achieved success in the composition of skills over a small number of objects. But their models do not consider the object's orientation - they assume a fixed orientation for the objects they are interacting with, which may not be true in real-world domains. 




We can see that the planning-based approaches, more specifically, PDDLStream based approaches, have the right framework for solving the problem we have defined. It allows us to to function in the hybrid discrete and continuous space, allows planning on models learned in the world, and composes them to achieve multi-step tasks. 

Hence, we propose to solve the problem of performing a complex multi-step task in a hybrid discrete and continuous environment by augmenting the PDDLStream framework. PDDLStream is a framework that itself augments Planning Domain Definition Language (PDDL) by adding conditional generators to sample values for continuous variables in the environment, thereby reducing the hybrid discrete and continuous planning problem to a discrete planning problem. We build upon this work by learning these conditional generators for sampling instead of hand-designed samplers defined in PDDLStream. We define affordance models as these learned conditional samplers that tell what actions are possible as well as how to perform them.They are learned for each action and object combination. For example, in the task where the agent has to move an object out of the way to reach a goal location, the affordance models generate pickable positions from where the agent can pick up the object and also generate the droppable location, from where the agent can drop the object in the desired location and finally be able to move to the goal location.

Our paper makes the following contribution :

\begin{itemize}
    \item We explore the idea of using affordance models to generate values for continuous variables as opposed to hand-designed samplers in the PDDLStream framework.
    \item  We show that learning affordance models at this level of abstraction helps in easy integration of these learned models, which in turn helps improve the overall system.
\end{itemize}


The rest of the paper is structured as follows. Section \ref{related_work} contains the related work. Section \ref{problem_formulation}, problem formulation gives a formal description of the problem. Section \ref{PDDL_with_affordance} contains our extension of PDDLStream using affordance learning to solve the problem described in section \ref{problem_formulation}. Section \ref{experiments_results} contains the experiments that were performed where we show the improvements by using affordance learning when compared to hand designed models. Section \ref{conclusion} contains the conclusions and section \ref{future_work} contains the future work.

\section{Related Work}
\label{related_work}

\textbf{Learning based approaches to perform tasks}: Existing works on embodied agents with pure learning-based approaches try to train reinforcement learning agents to do tasks through trial and error in the environment. They receive a large reward upon completing the tasks. Some of the popular reinforcement learning-based approaches use curiosity (\cite{curiosity_driven_exploration}) and coverage (\cite{Learning_Exploration_policies_for_Navigation}) to overcome the issue of sparse rewards. They are generally trained for very specific tasks such as exploring the world they are in or simple tasks like navigation to a point in the environment. Hence, they end up learning policies for exploration \cite{Learning_Exploration_policies_for_Navigation} or policies for specific tasks (\cite{Cognitive_Mapping_and_Planning_for_Visual_Navigation}). In our work, the learning is not done for any task but over properties of objects, and the tasks are completed by planning a sequence of actions based on the learned affordances. These affordances are not specific to any of the tasks.


\textbf{Affordance learning for tasks}: Some of the recent work in affordance learning in conjunction with reinforcement learning has been used to learn to perform tasks in 3D environments. In these approaches, the affordances models learn to predict possible actions in the given state of the world from the egocentric view the agent has. Now, an RL agent is then trained to perform tasks in the world by using this vision-based affordance model to narrow the set of possible actions in any given situation, which helps in learning to perform tasks faster.
( \cite{Learning_Affordance_Landscapes_Interaction_Exploration} \cite{Learning_About_Objects_by_Learning_to_Interact}). In our work, we learn to be capable of performing any task, even tasks not seen before but which are possible to execute through a composition of existing skills. The other approach to using affordances to perform tasks is to learn affordances not only to predict what actions are possible at a given moment but also what future actions are feasible if a certain action is taken (\cite{deep_affordance_foresight}). The main difference with our work is that they too limit their affordance to what actions are feasible whereas our affordances provide us the information about how to perform an action along with what actions are feasible. 

 
\textbf{Planning-based approaches to perform tasks}: PDDLStream, as described in the previous section, is a framework for solving TAMP problems that converts the discrete and continuous planning problem into a discrete planning problem using streams. Streams are conditional samplers and also have a declarative component that certifies the facts generated by the samplers about the properties the generated values satisfy in the given environment. These generated values are then used as facts in a discrete planning problem. Learning-based approaches which extend the PDDLStream framework (\cite{Learning_compositional_models}) learn the samplers to generate values for certain continuous variables. However, their models depend on the fact that the object orientation does not change. Our approach is an extension of PDDLStream work where we learn the conditional samplers as opposed to human-designed samplers, and the learned models are invariant to the object orientation/pose in the world. 




\section{Problem Formulation}
\label{problem_formulation}

Our objective is to solve the TAMP problem using planning algorithms.  That is, given a starting state and a goal state of the world, along with the information about the world, generate a sequence of actions in the world that enables us to go from the initial state to the goal state. 

We define our problem in the PDDLStream framework. This formulation helps us have a compact representation of the state of the world, actions, and the required goal. It also helps us use existing PDDL problem solvers to solve our TAMP problem in the hybrid continuous and discrete world. 


\subsection{Preliminaries}

First, we give a brief summary of PDDL and PDDLStream and then properties about the world we are operating in. Furthermore, we also give a formal description of our problem. We then give a brief overview of how a PDDLStream problem is solved. 

\subsubsection{Planning Domain Definition Language (PDDL)}
\label{PDDL}

PDDL has three main components - Predicates, states, and actions.  $(i)$ A predicate $p$ is a Boolean function. A fact $p(x)$ is a predicate $p$ applied on an object $x$ that evaluates to true. For example, a predicate $On (?body, ?region)$ checks if $?body$ is placed on $?region$. When the objects are instantiated by specific values, e.g. On( body1, region1), it evaluates to true if body1 is on region1 or false otherwise. This is known as a literal. $(ii)$ A $state$ $S$ is a set of literals.  $(iii)$ An action changes the state of the world. It has  has a parameter tuple $X = \langle x_1, x_2, ... x_k \rangle $ that is the set of objects in the world the action can affect. An action also has a set of preconditions, a list of literals that must be satisfied for an action to be applicable in any given state. Actions also have a set of effects which is also a list of literals.The effects define changes to the state of the world after applying the action in the world. For example, action $PickupObject (?obj, ?pose, ?position)$.It has as the set $(?obj, ?pose,?position)$ as parameters, which represent $object$, $object$ $pose$ and $agent$ $position$  respectively. It defines the ability of the agent to pick object $?obj$ at pose $?pose$ being at $?position$.The preconditions for this action are $(AtPose(?obj, ?pose)$ and $AtPosition(?position)$ checks if $?obj$ is at $?pose$ and if the agent is at $?position$ . The effects of this action are $(not$  $(AtPose(?obj,?pose))) and (PickedUp(?obj))$ which state that the object is no longer at $?pose$ and has been picked up by the agent.

The initial state of the world, $init$, is the starting state of the world and is a set of literals. $Goal$ is the goal state represented by a set of literals we want to be true in the world. $Actions$ is the set of all actions that are possible in the given environment. A PDDL problem is a tuple T = $ \langle Init,Actions, Goal \rangle $. This is a discrete PDDL problem if all the variables and action parameters are discrete-valued. The solution to the PDDL problem is a finite sequence of actions when applied in order from the start state,which will lead to a state with all literals in the goal state being true. There exist many popular search algorithms to solve discrete PDDL problems. In this work, we use the fast downward planner (\cite{fast_downward}). 

\subsubsection{PDDLStream}
\label{PDDLStream}

PDDLStream is a framework that extends PDDL by incorporating sampling procedures using streams. Streams are conditional generators with  procedural and declarative components. The sampling procedures are used to sample values for the continuous variables in the world. The declarative component defines the properties the inputs, and the outputs of the streams satisfy. The overall method PDDLStream uses to solve a given PDDL problem with continuous variables is as follows. Let $Search$ $(Init,Actions,goal)$ be an algorithm which can solve discrete PDDL problems.The hybrid discrete and continuous problem must be reduced to a discrete PDDL problem to utilize this $Search$ algorithm. This is done by sampling values for the continuous variables. These sampled variables are then added to the $Init$ state as facts based on the conditions they satisfy. The $Search$ algorithm then tries to find a sequence of actions using all the facts it has about the world to reach the foal state. If a solution is found, it returns the plan as the final solution. If not, streams are used to generate more facts about the world and this process is repeated until either a solution is found or a stopping condition such as timeout is hit. 



\subsubsection{Streams} 

 A stream is a conditional generator with a procedural component and a declarative specification of all the facts its inputs and outputs must satisfy. In this section, we go over each component of streams, and we also go more into detail as to how streams are used to generate values and how the generated values can be used as facts in the world.

A generator $g$ is a sequence of object tuples $y$. On call to the function $next(g)$, the next value in the sequence of tuples is returned if it exists. A conditional generator takes an input $x$ and produces an object tuple $y'$ related to the input $x$ if it exists. 

The declarative specification of the stream is represented through the domain and certified facts. Each stream has a domain, which is the set of predicates the inputs must satisfy. Each stream also has a set of certified facts, that is, all the outputs produced by the stream satisfy the predicates in the set of the certified facts. These certified facts that are sent back to the planner to consider as facts in the world while solving the planning problem. 

The procedural component of streams is what defines how to actually generate the values which satisfy the required constraints. This is generally a programmatic implementation of a function that takes the domain facts as input, also produces outputs based on the constraints in the certified facts they must satisfy. The implementation of the stream is considered a black box for the planner as it only utilizes the produced outputs and the conditions they satisfy. Hence, the sampling and the planning are two separate modular components. We will exploit this feature to change the process of sampling without changing the planning algorithm. 

A stream instance $s(x)$ is a stream where its input parameters have been initialized by object tuple $x$.

A PDDLStream problem can be represented as a tuple $T = \langle Init, Actions, Streams, Goal \rangle $ where $Streams$ is the set of streams defined for the domain we are operating in. 


\subsection{Problem Definition}

We now define our TAMP problem in the PDDLStream formulation. We first define what the initial state and goal state is in our environment, then the available actions, and then what streams are required for our environment.

\subsubsection{Init and Goal State}

\begin{figure}[h]
\begin{center}
\fbox{\centering\includegraphics[width=3in]{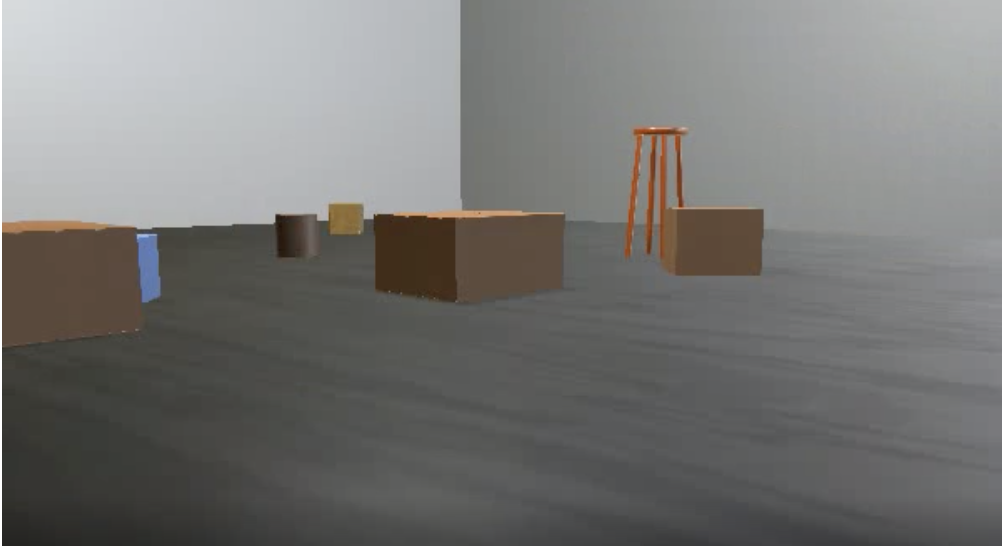}}\par
  \caption{Example environment where a task has to be performed}  
  \label{generic_AI2Thor_sce}
\end{center}
\end{figure}

As seen in fig \ref{generic_AI2Thor_sce}, 
the agent is in the AI2Thor environment, a virtual 3D environment \cite{ai2thor}. The environment contains a set of objects in a room. The input to the agent is the bounding box information of all the objects in the environment. Using this information, we create our $\hat{I} = init$ state of the world, that is, create predicates for objects and their orientation/poses in the room. This $init$ state contains information about all the objects of the world and the agent's own information. The goal $\hat{G}$ is described as a set of literals, similar to the initial state. For example, if the goal is to pick up a cube, the goal state would be $PickedUp(cube)$. \\
We assume that we are working in a fully observable setting where the information of all the objects is available to us in the very beginning. We also assume we have a perfect perception system, and hence we have the object orientation and bounding box information.

\subsubsection{Actions}
\label{actions}

AI2Thor has a host of actions that enable a large number of interactions with the objects in the environment as well as navigation. In our work, we consider the following subset of actions :$\hat{A} =$ $MoveAgent$, $OpenObject, PickupObject$, $CloseObj$. A solution to the PDDLStream problem is a sequence of these actions and values of the parameters that satisfy the preconditions (the values for the parameters are returned as part of actions to execute). An example of a full description of an action is as follows :

\begin{verbatim}
   (:action PickupOobject
    :parameters (?o ?p ?q)
    :precondition (and
                    (PickablePos ?o ?q ?p)
                    (AtPose ?o ?p)
                    (AtPosition ?q))
                  )
    :effect (and (PickedUp ?o) (not (AtPose ?o ?p)) )
 )
\end{verbatim}

In this example of the $PickupObject$, as mentioned in section \ref{PDDL},
$?o$ stands for the object to be picked up, $?p$ stands for the pose of the object, $q$ stands for the position of the agent from where the object is to be picked up. The preconditions are that the object is at that expected pose, the agent is at the expected position, and the position is a $Pickupable$ position. It is the $pickupable$ position that the streams generate. A stream is defined for generating the continuous variable $position$, which satisfies the constraint that it is a $pickable$ position. 

\subsubsection{Streams}

Streams are required to generate values for the continuous variables $position$, $object$ $pose$. The number of streams we need is a lot higher as we need to generate these continuous-valued variables with different certifications - we need to generate agent positions for opening and closing an object, and so on. Let the set of all the streams required be $\hat{S}$. An example of a stream is the $sample-pickable-point$ stream. 

\begin{verbatim}
  (:stream sample-pickable-position
    :inputs (?o ?p)
    :domain (and (Pose ?o ?p) (Pickable ?o))
    :outputs (?pos)
    :certified (PickablePos ?o ?q ?pos)
  )
\end{verbatim}
  
  This stream takes in an object and its pose information as input. It checks if the object is a pickable object and that specified pose is the pose of the object. It then generates a position $?pos$, which it certifies as being a $PickablePos$ (A position from which the object is pickable). This generated value and the certified fact are then used to satisfy the preconditions of the action $PickupOobject$ mentioned above.

\subsubsection{Formal definition of our TAMP problem in PDDLStream}

We have defined the initial state, actions, and streams for our problem formulation. Hence, the PDDLStream formulation of our problem definition is as follows: $T= \langle \hat{I}, \hat{A}, \hat{S}, \hat{G}\rangle$ and the solution is a set of actions that when applied sequentially, change the state of the world from initial state $\hat{I}$ to goal state $\hat{G}$.
The environment dynamics captured through the actions are described in the $domain.pddl$ file in the PDDL syntax. The specific instance of the problem we are solving, a particular arrangement of objects and goals, is described in the $instance.pddl$ file in PDDL syntax. The declarative part of the streams is also defined in the PDDL syntax in a $stream.pddl$ file. 

\subsection{Solving a PDDLStream problem}

We have defined the TAMP problem as an instance of a problem in the PDDLStream framework. In this section, we cover how to solve problems defined in the PDDLStream framework. 

The existing PDDLStream framework takes the four tuple as input and outputs a sequence of actions. There are multiple algorithms that can be used to solve the PDDLStream problem. However, we present only the simplified version of the incremental algorithm from \cite{PDDLStream_main} (Other algorithms and details of this algorithm are omitted as our proposed approach works with any algorithm and does not depend on any other part of the algorithm). 

\begin{algorithm}[H]
\SetAlgoLined
 Stream instance generation\;
 \While{True}{
  $\hat{I}$ = APPLY-STREAMS($\hat{S}, \hat{I}$)\;
  Plan = Search($\hat{A}, \hat{I}, \hat{G}$) \;
  \If{Plan != None}{
   return Plan\;
   }
 }
 \caption{Incremental algorithm ($\hat{A}, \hat{I}, \hat{S}, \hat{G}$)}
\end{algorithm}

The stream instance generation is a process where the facts given in the initial states are used to generate stream instances. That is, all the facts that satisfy the domain constraints of a stream are used to create a stream instance. The APPLY-STREAMS function evaluates the stream instances and samples values for the continuous variables and adds these new generated facts from streams to $\hat{I}$. The $SEARCH$ is then run for searching a plan from the given facts. If it succeeds, a plan is returned, else more stream instances are generated using the newly generated facts, and these new stream instances are evaluated to generate more facts, and $SEARCH$ is run on this larger set of facts. This process is repeated until a plan is found or a stopping condition is met, such as exceeding the time limit.  

\section{PDDLStream with affordances}
\label{PDDL_with_affordance}

This section will cover our proposed addition to the existing PDDLStream framework and how to incorporate this new method to improve the existing framework.


\subsection{Affordances: Motivation and definition}

We now know that streams form an integral part of solving a PDDLStream problem. However, these streams have a few drawbacks. Each of these streams, in the original PDDLStream framework, are hand-designed. The drawbacks of defining streams this way are multi-fold. First, since we are designing the samplers based on our knowledge of the world, there might be information about the world we are missing and leading it to to generate values that might not be useful. Second, since streams can be used in more complex situations such as generating droppable configurations for objects, it might not be possible to hand-design these samplers (droppable position is more complex than pickable position because the drop action in AI2Thor is less reliable with where it drops the objects). 


Hence, we propose learning these samplers as generative functions. The main intuition behind our idea is that each of the continuous variables we are trying to sample values comes from a particular distribution that depends on the object itself and the action for which the value is being sampled. 

For example, in the $Sample-Pickable-Position(cube, pose)$ stream instance, we try to sample for the continuous variable position from where the object cube can be picked up. We can see that it also depends on which object we are sampling this position for.  Hence, we conclude that a. distribution exists for the continuous variable position over each action on a particular object. We now try to learn a generative function for this distribution that can generate a value from this distribution whenever necessary. If we can learn to generate these values using this generative function, we can guarantee that the generated values have the required properties. In this case it would be that the generated position is indeed a position from which the cube can be picked up. 
We learn a separate probability density function for each object and every actionable property possible in the environment. This is required as each object has its own distribution of values for which an action will be successful. 

 We use the Kernel density estimation (KDE) method to learn the required distribution. Learning to sample solves both the aforementioned problems about streams. Learning ensures that we do not miss any information which can be missed while hand designing, and also, when some functions are too complicated to hand design, the function can be learned efficiently.


\subsection{Learning the affordances}

Our approach is to first try to perform actions with an object in an empty environment. We let the agent interact with the object by attempting to perform all the actions in the environment. The data is collected the data for the runs where the action was successful. We then use this data and learn the distributions of the continuous variables required for a particular object for different actions. The system learns separate density function for each of the action-object pair. We repeat this process for each object. The learned generative density functions are what we call affordances. These affordances tell us what actions are possible with what objects (if none of the actions were successful while training, we assume that particular action is impossible for the object of interest in that run). The affordances also tell how to perform the action by generating values for the continuous variables that satisfy the action's preconditions. Hence, these learned affordances become our conditional samplers which guarantee that when they generate a value, it satisfies the required declarative constraints defined as part of the stream definition. 

This learning can also be done online - that is, when the agent is put into a new environment, it can go about trying to interact with the objects in the environment. As it succeeds with interactions, it can learn from that data after a certain number of successful interactions of each type with every object. 


\section{Experiments and Results}
\label{experiments_results}

We use our approach to solve three different tasks with increasing order of complexity. We conduct all the experiments in the AI2Thor simulator.
It is a 3D virtual environment with a physics engine that helps in simulating real-world tasks.  We can generate environments with different configurations with any number of objects and have a large number of interactions with these objects in the environment. 
We compare the two ways of solving the problem - one with hand-designed streams and one with learned streams to generate values for continuous variables. We compare these approaches by comparing the success rate of the approaches while performing these tasks. We also see the effect of the number of objects in the room on the same parameters for both approaches. The first step in our process of generating plans for solving tasks is to learn about the objects in the training phase. We  use these learned models while solving the tasks and compare the results to the system performs with hand-designed models.

The complexity of a task is defined as the number of required actions to go from the start state to the goal state. 

\subsection{Tasks}

\subsubsection{Task 1:Move object from 1 location to another}


The first task for the agent is to pick up an object and place it in a different location. In this task, the streams of importance are the streams that generate the positions from where an object is pickable and also from what position an agent should be dropped from to place it in the required destination location. The complexity of the task is four, as the expected solution has four actions. An example of this task can be seen in Figure \ref{generic_AI2Thor_sce}, where the agent is expected to pick the blue object and move it to a different location.

\subsubsection{Task 2 : Pickup object while being completely blocked initially}

\begin{figure}[h]
\begin{center}
\fbox{\centering\includegraphics[width=2.5in]{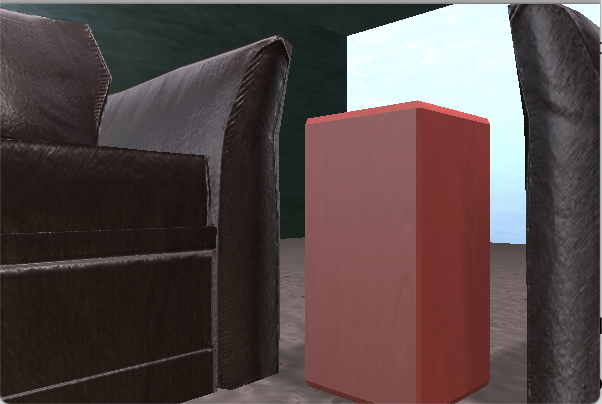}}\par
  \caption{Example environment for Task 2 }  
  \label{task_2_scene}
\end{center}
\end{figure}

In this task, the agent is surrounded by large unmovable objects and one small movable object. The small object can be moved out of its position to reach the required goal location. The complexity of the task is six, as the expected solution has six actions. An example of this task can be seen in Figure \ref{task_2_scene}, where the agent is expected to pick the red cube to be able to move to a region currently blocked by sofas and the red cube.


\subsubsection{Task 3 : Place an object in the world at a certain height and construct a support structure to hold it}

In this task, the goal is specified as a mid-air pose for the object. The agent is expected to create a support structure by moving other objects underneath the mid-air goal pose of the first object.  The complexity of the task is eight.

\subsection{Experiments}

For our experiments, we create environments with random object positions and poses. For all the tasks, we test them in $2$ separate environment configurations. First, only the objects which are relevant to the task are in the room. For this, we generate scenes with only the objects necessary for completing the task. Next, we add $20$ objects to the scene which are not useful to the task but exist in the room. We generate scenes for this set of experiments by randomly placing 20 objects in the room. This is to test how the number of objects in the room affects the task at hand.  We run the experiment 10 times for both settings.

\subsection{Results}
\label{results}


\begin{table}[]
\centering
\caption{Task Success Rate (\%) }
\label{result_table}
\begin{tabular}{@{}cccc@{}}
\toprule
\textbf{\begin{tabular}[c]{@{}c@{}}Number of extra \\ objects in scene\end{tabular}} & \textbf{Task 1} & \textbf{Task 2} & \textbf{Task 3} \\ \midrule
0 (PDDLStream)    & 70  & 70 & 10 \\
0 (Our approach)  & 100 & 80 & 40 \\
20 (PDDLStream)   & 70  & 40 & 10 \\
20 (Our approach) & 100 & 70 & 30 \\ \bottomrule
\end{tabular}
\end{table}
We can see from Table \ref{result_table} that using the learned affordance models helps us achieve a higher success rate in task completion. The majority of the failures for both approaches came while trying to drop an object. While our approach can achieve a fair amount of success in all the tasks, the vanilla PDDLStream approach with hand-designed models fails as the object is being dropped incorrectly in a majority of the tasks. 

The complexity of the task itself does affect the overall performance - we see a performance decline in both approaches as the tasks get more complex. As the length of the task increases, the chances of failure in any one of the steps increases, thereby increasing the overall chances of failure. The performance on task 3 is particularly poor as dropping objects on top of each other as it is a hard task, and very accurate values must be sampled for this to be successful. 

Adding extra objects in the room does affect the success rate but not too much. They come into play when they are very close to the goal location or objects of interest. If they are too close to the object of interest, the generated samples must be close to perfect to perform the actions. In this case, the learned models do better as they produce more accurate values for performing actions. 

\section{Conclusions}
\label{conclusion}

In this work, we describe a new way of incorporating learned models of the world into a planning framework for completing complex multi-step tasks in a hybrid discrete and continuous environment. We do this by extending the PDDLStream framework to include learned affordance models  of the environment. The affordance models are learned for each object and action in the environment. They are used to sample values for continuous variables in the world. This helps reduce the hybrid continuous and discrete world into a discrete planning problem that we solve using a symbolic planner. Through our experiments in the AI2Thor simulation environment, we show that this extension leads to the agent being able to solve tasks more consistently in complex environments. Having the capability to plan over learned affordances helps in achieving complex tasks very easily the through composition of these learned models. This framework of learning and planning can be used in a hybrid environment to learn and execute tasks faster than existing approaches for the same. 


\section{Future Work}
\label{future_work}

There are few potential ways to extend the work described in our paper. 

First, we can extend the problem statement to cover a more generic view of the world and solve tasks in more complex environments. In our current work, planning happens in a fully observable world. It is possible that the world is partially observable, which it is in many cases (such as an object being stored inside a container and the agent only knows it's in one of the many containers but is not sure which one) and hence planning over belief states might be necessary in these cases. Our framework allows us to easily extend our work to tackle this broader class of problems.

The other improvement to our described work is the choice of the generative function itself. In the current approach, we learn a separate model for each object-action pair. We can try to learn a more generic model over actions or over a set of objects. This will reduce the amount of learning required and can also lead to better generalization to objects not seen before. 

Lastly, improvements can be made to the PDDLStream algorithms to increase efficiency further. We can see that stream instance generation can be improved to generate a smaller number of stream instances. Currently, the algorithm exhaustively generates stream instances for all the facts that satisfy the domain of the defined streams. This leads to a large number of stream instances being generated which are not useful to the task and increase the planning time as there is a larger set of facts that need to be searched to find a plan. This efficiency can further be improved by looking at the stream instances which have been generated. There are some stream instances that are more useful to the planner as compared to a few others. Designing a way to rank the generated stream instances using heuristics will help use the more useful ones and discard the less useful ones. 



\bibliography{iclr2021_conference}
\bibliographystyle{iclr2021_conference}


\end{document}